# Boosting SISSO Performance on Small Sample Datasets by Using Random Forests Prescreening for Complex Feature Selection

Xiaolin Jiang[1], Guanqi Liu[1], Jiaying Xie[1], Zhenpeng Hu[1*]

[1]School of Physics, Nankai University, Tianjin, 300071, China.

*Corresponding author E-mail(s): zphu@nankai.edu.cn

## ABSTRACT

In materials science, data-driven methods accelerate material discovery and optimization while reducing costs and improving success rates. Symbolic regression is a key to extracting material descriptors from large datasets, in particular the Sure Independence Screening and Sparsifying Operator (SISSO) method. While SISSO needs to store the entire expression space to impose heavy memory demands, it limits the performance in complex problems. To address this issue, we propose a RF-SISSO algorithm by combining Random Forests (RF) with SISSO. In this algorithm, the Random Forest algorithm is used for prescreening, capturing non-linear relationships and improving feature selection, which may enhance the quality of the input data and boost the accuracy and efficiency on regression and classification tasks. For a testing on the SISSO's verification problem for 299 materials, RF-SISSO demonstrates its robust performance and high accuracy. RF-SISSO can maintain the testing accuracy above 0.9 across all four training sample sizes and significantly enhancing regression efficiency, especially in training subsets with smaller sample sizes. For the training subset with 45 samples, the efficiency of RF-SISSO was 265 times higher than that of original SISSO. As collecting large datasets would be both costly and time-consuming in the practical experiments, it is thus believed that RF-SISSO may benefit scientific researches by offering a high predicting accuracy with limited data efficiently.

## 1 INTRODUCTION

Data-driven approaches in materials science significantly accelerate material discovery and optimization, reducing costs and enhancing the success rate of material development. [1-5] Symbolic regression is an effective data-driven modeling technique that automatically discovers mathematical expressions from data, representing relationships between variables. [6-12].Among the numerous symbolic regression algorithms, the Sure Independence Screening and Sparsifying Operator (SISSO) [13] introduced by Ouyang et al. has attracted people's attention. It can extract material descriptors from large datasets, which is applicable across various fields of study. [8,14-15] It generates a large feature space from the original space, and then selecting features from this new space to build models with compressed sensing algorithm.

Consequently, SISSO's requirement to create and partially store the entire expression space leads to exponentially growing memory demands with increasing features and complexity, making it resource-intensive for complex problems. To address this issue, Ouyang et al. developed VS-SISSO [16], which combines symbolic regression with iterative variable selection (random search) [17-18] to optimize the model with numerous input features. [19-21]. Alternatively, Xu et al. proposed i-SISSO [22], which integrates mutual information (MI) [23] and minimum redundancy maximum relevance (mRMR) algorithms [24] to optimize feature combinations for maximum relevance and minimum redundancy. Obviously, the number of input features directly affects the performance of SISSO, VS-SISSO, or i-SISSO, as all input features are considered in the SO process. Therefore, we propose a more straightforward idea that performing a prescreening to get the important features for an effective input of SISSO may reduce the computational complexity and storage costs, and save time. The decision tree model [25] can describe nonlinear relationships efficiency for datasets of different sizes, which may help us to realize the above idea. Since a single tree model may result in high variability, it is believed the Random Forest algorithm [26-27] should be a better choice for the prescreening of features.

On the other hand, it would be expensive and time-intensive in the experiments to gather extensive datasets, especially for physics, chemistry, material, and life science. The lack of abundant data may reduce the performance of machine learning algorithms for those areas, in particular exploring new phenomena. Random Forests generate multiple tree models based on subsets obtained from bootstrapped samples, then vote on data importance. This resampling method effectively enlarges datasets, which is naturally friendly to the scientific researches with limited data. It would be expected to combine Random Forests with SISSO for those researches.

Herein, we combine Random Forests with SISSO to describe certain nonlinear relationships, resulting in RF-SISSO. Taking the SISSO's verification problem for 299 materials [13,28] as an example, RF-SISSO and SISSO were compared on training datasets of various sizes. The training sample sizes of 224, 150, 75, and 45 were randomly selected from the dataset of 299 materials, while the other samples' data were used for testing in each case. For five parallel testing on each training sample size cases, RF-SISSO maintained a predicting accuracy above 0.9 in all cases, whereas SISSO's accuracy was below 0.9 in the 45-sample subsets. As the prescreening by Random Forests effectively reduce the number of input features, RF-SISSO's descriptor regression efficiency was higher than that of original SISSO in all cases, notably reducing time costs. Meanwhile, the RF-SISSO algorithm were also compared with i-SISSO algorithm [22], where RF-SISSO demonstrated higher regression efficiency than that of i-SISSO with a comparable accuracy.

## 2 METHOD

### 2.1 Experimental Dataset and Environments

SISSO and RF-SISSO were tested on the $A_xB_y$ classification problem of metal/non-metal binary materials using experimental data collected by the authors who developed SISSO. [13,28] The data sources included the WebElements (atomic) and SpringerMatters (structural) databases.

**The features analyzed were:**

- Pauling electronegativity ($\chi$): $\chi_A, \chi_B$
- Ionization energy ($IE$): $IE_A, IE_B$
- Covalent radius ($rcov$): $rcov_A, rcov_B$
- Electron affinity ($EA$): $EA_A, EA_B$
- Number of valence electrons ($v$): $v_A, v_B$
- Coordination number ($CN$): $CN_A, CN_B$
- Interatomic distance: $d_{AB}$
- Atomic composition: $nA$ ($nB = 1 - nA$)
- Packing fraction: $V(V_{atom}/V_{cell}$, where $V_{atom} = 4\pi(rcov)^3/3)$

Combining data from WebElements and SpringerMatters resulted in 15 prototypes, covering a total of 299 materials and 16 features. The testing was conducted on hardware configured with an Intel(R) Xeon(R) Gold 5218CPU @ 2.30 GHz, 128 GB memory, and running on 4 cores.

**2.2 Experimental Process**

**(1) Feature Evaluation using Random Forests:**

- The features of all materials were evaluated using the Random Forests algorithm. The importance of each feature was determined using the Gini coefficient.

- Due to the small sample dataset, both sample division and Random Forest selection were randomized, which led to varying prediction accuracy for each cross-validation and potential experimental contingencies. To mitigate this, feature assessment was repeated at least 50 times using the Random Forests algorithm.

- The importance scores for each feature were obtained and ranked. These scores, once normalized, allowed for a clear comparison of feature importance. The features, those with an importance score of 5 or higher, were selected for the SISSO calculation.

**(2) Operator Regression using SISSO:**

- After identifying important features with Random Forests, SISSO was employed to filter key features further. This process yielded two-dimensional descriptors for classification through symbolic regression.

- The total sample set was divided into five subsets, with samples sizes of 224, 150, 75, and 45. Each subset was randomly selected five times. The flowchart of the RF-SISSO algorithm is shown in Fig.1.

- RF-SISSO selects the top 8 features with importance scores of 5 or higher following the feature ranking by Random Forest. SISSO, without the integration of Random Forest, continues to be represented by the original 16 features.

- The accuracy of the descriptors obtained from original SISSO and RF-SISSO regressions were evaluated using the Support Vector Classifier (SVC). [29]

For detailed principles of algorithm application, please refer to the supporting information and original literature.[13,26-27]

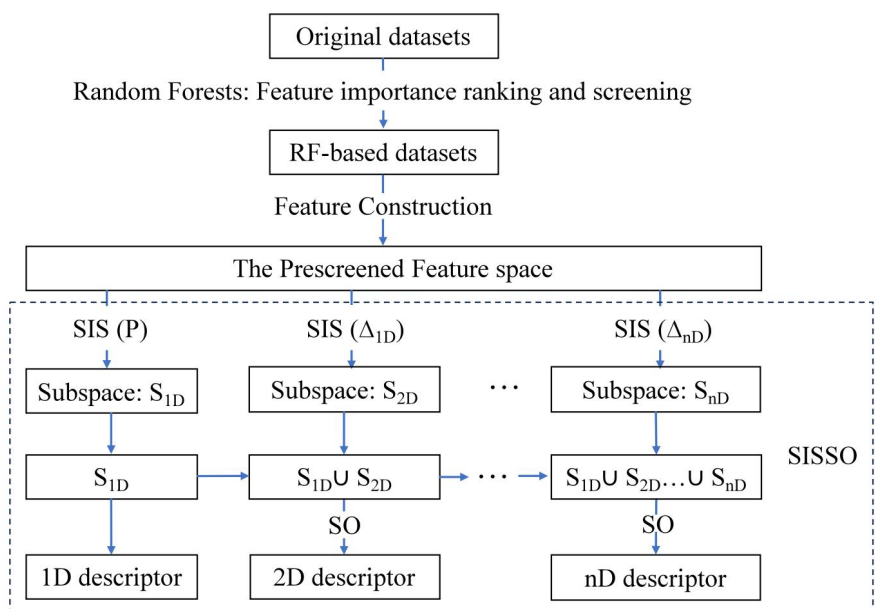

Fig.1 Flowchart of RF-SISSO algorithm

## 3 RESULT

### 3.1 Random Forests Algorithm Feature Screening and Performance of SISSO (SISSO & RF-SISSO) in datasets of varying sizes

The study evaluated the importance of features using the Random Forests algorithm. Fig.2(a) shows a histogram of the features affecting metal/nonmetal properties as determined by the Random Forests analysis. The top eight important features with feature importance ratings above 5 were identified: the electronegativity of the "A" atom and "B" atom ($\chi_A$ and, $\chi_B$), packing fraction (V), valence electron number of the "A" atom and "B" atom ($v_A$ and $v_B$), electron affinity energy of the "A" atom and "B" atom ($EA_A$ and $EA_B$), and ionization energy of the "B" atom ($IE_B$).

To compare the accuracies of models from SISSO and RF-SISSO, we used the obtained 2D descriptors as training features to classify the metal/nonmetal materials using the SVC classifier and recorded their classification accuracies. For a simple presentation, the accuracy of SISSO/RF-SISSO is used to represent the accuracy of 2D descriptors from SISSO/RF-SISSO. As shown in Fig.2(b), the accuracies of both SISSO and RF-SISSO are above 0.9 with the 224-sample subset, where SISSO generally performs slightly better. However, RF-SISSO's accuracy tends to increase as the dataset size decreases. With the 150-sample subset, RF-SISSO outperformed SISSO in two training sets and matched SISSO in one testing set. With 75-sample subset, RF-SISSO had one training set significantly higher than SISSO and outperformed SISSO in three testing sets. For the 45-sample subset, RF-SISSO consistently outperformed SISSO, with all training accuracies significantly higher and only one testing set lower. In this case, two training sets of SISSO have accuracy lower than 0.9, which demonstrates RF-SISSO's advantage in achieving high accuracy with smaller datasets. Here, RF-SISSO can achieve higher accuracy than SISSO is mainly attributed to RF's ability to capture complex nonlinear relationships and interactions between features and target properties. By identifying and removing irrelevant or redundant features with RF prescreening, the efficiently reduced feature space allows SISSO to operate more efficiently and focus on the most informative features. It helps to mitigate the risk of overfitting, as the model is less likely to learn noise from irrelevant features.

The major advantage of RF-SISSO over SISSO is its shorter operator regression time and higher training efficiency. Fig. 2(c) and (d) show that the average operator regression time of SISSO is about 26 times longer than that of RF-SISSO for the 224-sample and 150-sample subsets, 39 times longer for the 75-sample subset, and up to 265 times longer for the 45-sample subset. This demonstrates that by using RF to eliminate redundant features, the regression efficiency of SISSO is effectively improved. In addition, we used the Random Forests algorithm alone to classify the data. As shown in Table S1, the testing accuracy of RF is not as good as SISSO, but the computing time is much shorter. The complementarity between RF and SISSO make RF-SISSO to have a more efficient processing of high-dimensional datasets to construct well-generalized accurate models.

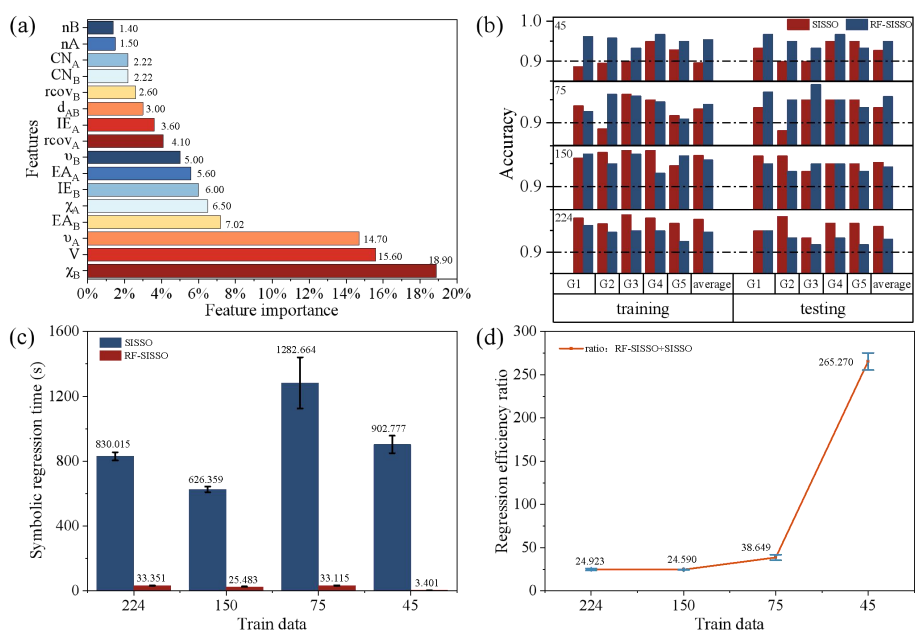

Fig.2 (a) Histogram of feature importance assessment, (b) Accuracy of training and testing of 2D descriptors for datasets of various sizes, (c) Mean regression time for SISSO vs. RF-SISSO trained operators (4 cpu cores), and (d) Training efficiency ratio of SISSO vs. RF-SISSO (4 cpu cores).

The two-dimensional descriptors obtained by symbolic regression of SISSO and RF-SISSO under different dataset scales were presented in Table 1. The descriptors were derived from the set of data with the highest accuracy in the five parallel sets for each sample size. Comparing the descriptors of SISSO and RF-SISSO across different data scales reveals that the complexity of RF-SISSO descriptors is consistently lower than that of SISSO. Notably, the selected features may vary in each experiment due to differences in the training set, even when the number of initial features is not limited in SISSO. The features in the RF-SISSO descriptor are similar to those of SISSO but more meaningful in physics, where the combinations like $V/\chi_A$ and $\chi_B/\chi_A$ recurs. It indicates that stronger electronegativity leads to increased exclusivity rather than sharing of charge, making it easier to form non-metals. Additional descriptors are available in the supporting information (Table S2-S5).

Table 1 Two-dimensional descriptor derived by symbolic regression using SISSO and RF-SISSO across datasets of different subsets.

| Datasets | | SISSO | RF-SISSO |
|---|---|---|---|
| 224 | d1 | $\chi_A/(V \times \chi_B) + \exp(-CN_B/nB)$ | $(V \times IE_B)^2 \times \exp(EA_A - \chi_A)$ |
|  | d2 | $|\chi_B/(V \times IE_B)| - nB \times |nA - nB|$ | $|1 - \chi_B/\chi_A - IE_B/(EA_A - \chi_B)|$ |
| 150 | d1 | $d_{AB} \times (V + nB) \times (\chi_A/nA - IE_B)$ | $(IE_B - 2\chi_B)/(\chi_A/V)^2$ |
|  | d2 | $(IE_B \times \chi_B \times (d_{AB} + rcov_A))/(\chi_A/V)$ | $|(\chi_B - EA_A) - |(EA_A + EA_B) - |EA_B - \chi_B|||$ |
| 75 | d1 | $(\chi_A/nA - IE_B \times V \times (d_{AB} + rcov_A)$ | $(V/\chi_A)^2 \times \chi_B \times (EA_A - IE_B)$ |
|  | d2 | $(V - nA) \times CN_B \times IE_B \times \chi_B/\chi_A$ | $\exp(IE_B)/(\chi_A/EA_B - \chi_B/\chi_A)$ |
| 45 | d1 | $V^2 \times \chi_B \times (d_{AB} \times IE_B)^3$ | $(V/\chi_A)^3 \times (IE_B + \chi_B)^3$ |
|  | d2 | $V^2 \times IE_B \times (d_{AB} \times IE_B)^3$ | $(V/\chi_A)^3 \times (EA_A + \chi_B)^3$ |

The results of metal/non-metal classification using the SVC classifier with the 2D descriptors obtained from SISSO/RF-SISSO symbolic regression are shown in Fig.3. Blue dots represent metals, red dots non-metals, and the yellow line the decision boundary. The classification for SISSO training and testing on one dataset of 45-sample subset are shown in Fig. 3(a) and (b), while the same for RF-SISSO are shown in Fig. 3(c) and (d). The RF-SISSO's classification boundary (Fig.3(d) and its inset) is notably clearer than that of SISSO (Fig. 3(b) and its inset). (See supporting information for other dataset's visualizations: Fig.S2-S4).

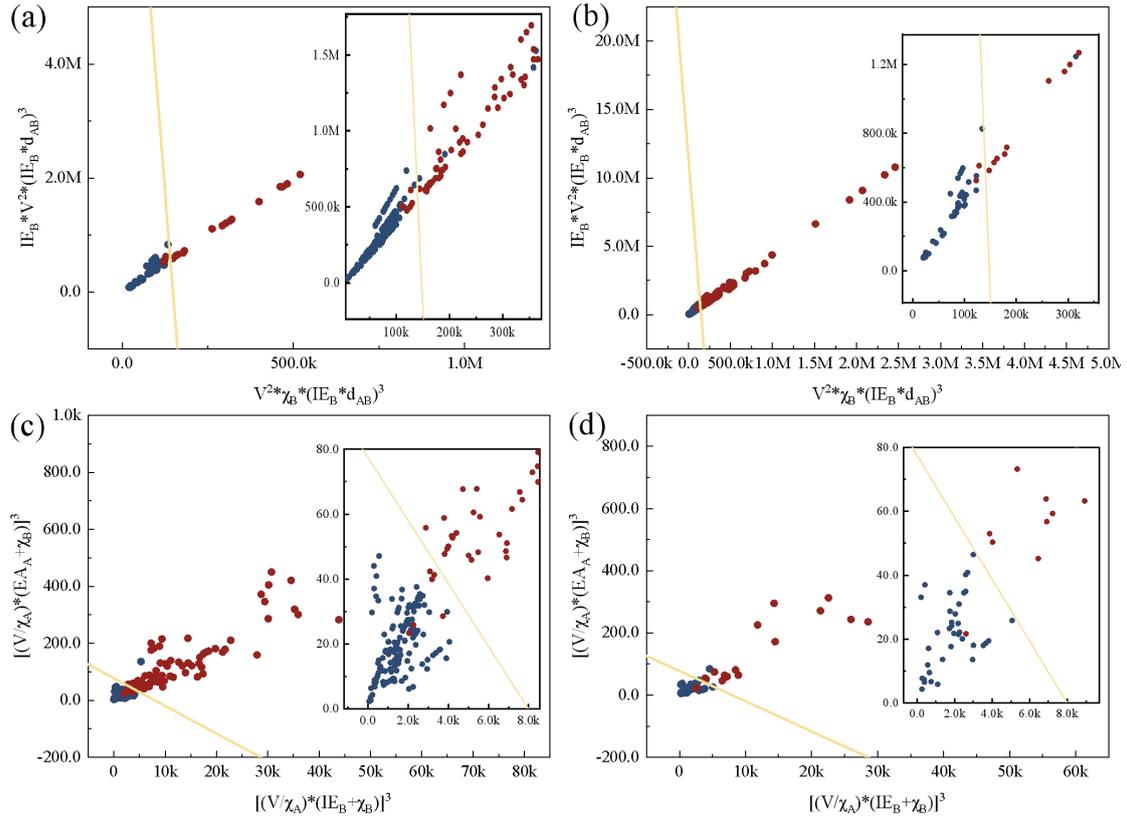

Fig.3 SVC classification visualizations for 2D descriptors obtained from SISSO and RF-SISSO in a 45-sample subset. (a) training and (b) testing sets with operators from SISSO; (c) training and (d) testing sets with operators from RF-SISSO.

For a comparison, we selected the top 6 and top 10 features as datasets and repeated the experiment. With a 45-sample subset (Fig.4(a)), the testing accuracy is the highest with the 8-

feature's regression. With 10 features, accuracies of three training and one testing sets are below 0.9, which indicates more features are not necessarily better. For the 6 features' case, accuracies of two training and one testing sets are below 0.9, showing that fewer features are also not optimal. Additionally, Fig.4(b) shows that operator regression time is the shortest with 8 features, indicating a maximum efficiency.

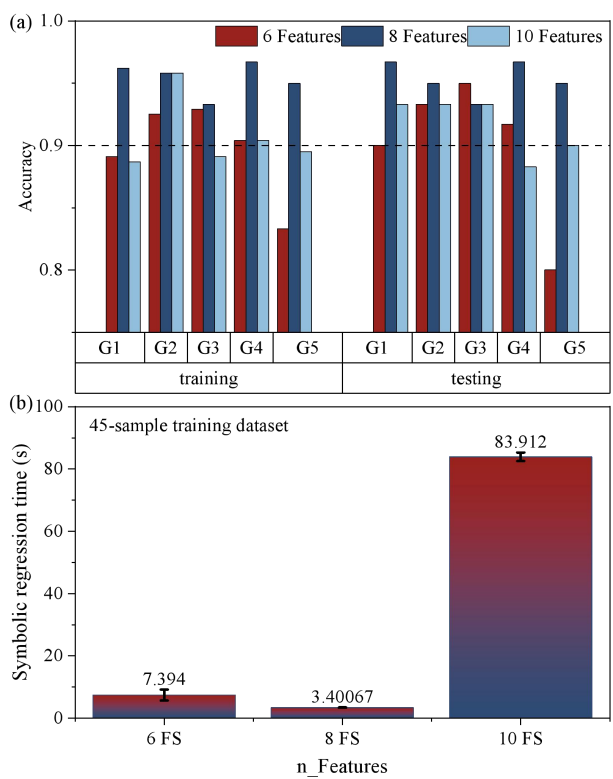

Fig.4 Comparison of RF-SISSO: (a) accuracy, and (b) operator regression time for a 45-sample subset with varying quantitative features.

Recently, Chong et al. [30] had established interpretable spectral-property relationships with the SISSO algorithm on small datasets. It is valuable to see whether RF-SISSO can make an improvement in regression efficiency and accuracy for these datasets. As shown in Table S6, operating the data of small datasets with sample sizes of 20 and 40, the results of RF-SISSO were improved in most cases. By removing redundant features, the efficiency of regression was significantly enhanced. (Fig. S5) It is clear that RF-SISSO does have the advantage than the original SISSO, particularly the efficiency.

Xu et al. [22] had proposed an improved SISSO (i-SISSO) by integrating MI and mRMR algorithms to address SISSO's limitations in high-dimensional model generation. As shown in Fig. 5, we have also compared i-SISSO and RF-SISSO with the same parameters on seven raw datasets from Xu et al. [22] The RMSE of different systems (Fig. 5(a)) are almost identical for the two methods, while the regression times show significant difference (Fig. 5(b)). As regression times of some systems are too short to be clearly observed, the ratios of regression times are presented in the inset of Fig. 5(b) for an easier comparison. Obviously, RF-SISSO is once more advancing in terms of regression efficiency while the difference in accuracy is neglectable.

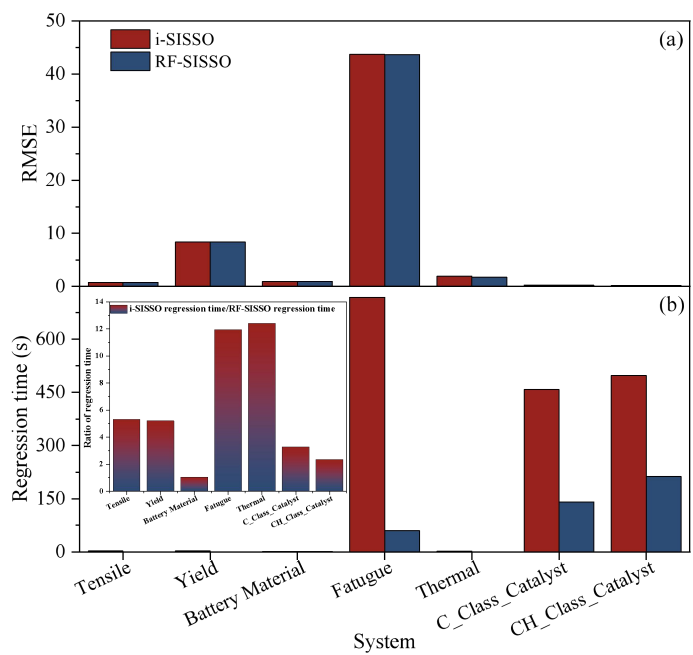

Fig. 5 Comparison of the i-SISSO and RF-SISSO algorithms: (a) accuracy, (b) operator regression time (inset figure shows the ratio of regression time).

**4 CONCLUSION**

In conclusion, our study demonstrates the improvements achieved by integrating the Random Forests algorithm with SISSO, resulting in the enhanced RF-SISSO method. This combination enables SISSO to maintain high-precision predictions even with small datasets. RF-SISSO not only improves prediction accuracy across various dataset sizes but also produces more concise 2D descriptors. Furthermore, RF-SISSO enhances regression efficiency, being up to 265 times faster than SISSO for the smallest subset. These findings highlight the robustness and efficiency of RF-SISSO, making it a valuable tool for material descriptor regression across diverse dataset sizes. The complementarity of the Random Forests algorithm and the SISSO algorithm can capture non-linear relationships more effectively, resulting in more accurate predictive models with good generalization ability. This approach also reduces computational complexity and improves operational efficiency.

**Acknowledgements** This work was supported by the National Natural Science Foundation of China (No. 21933006, 21773124), the Fundamental Research Funds for the Central Universities Nankai University (No. 010-63243091, 63233001) and the Supercomputing Center of Nankai University (NKSC).

**Supporting Information**

**1 Random Forests algorithm for feature screening**

Gives a set of material datasets $D = \{f_1, P_1\}, ..., \{f_N, P_N\}$, where $f_1, ..., f_N \in \mathbb{R}^M, P_1 ... P_N \in \mathbb{R}$, that is N materials are provided, each of which has M features. The first screening of features is undertaken using random forest. Random Forests [1] is a machine learning algorithm that integrates multiple trees with the idea of integrated learning, where the basic unit is the decision tree. Simply compare how much each feature contributes to each decision tree in the random forest, the larger the contribution, the greater the impact it has on the outcome. The contribution size is measured by the Gini index and the out-of-bag data error rate. The Gini value of the dataset D can be defined as:

$$Gini(D) = \sum_{k=1}^{k} \sum_{k' \neq k} p_k p_{k'} = 1 - \sum_{k=1}^{k} p_k^2 \qquad (1)$$

where assume that the value space of $P$ is $\{s_1, ..., s_k\}$, pk denotes the probability that $P$ takes $s_k$. The larger the $Gini(D)$, the greater the uncertainty of dataset $D$.

The Gini index corresponding to characteristic $F_m$ be defined as:

$$Gini\_index(D, F_m) = \sum_{v=1}^{V} \frac{|D^v|}{|D|} Gini(D^v) \qquad (2)$$

where feature $F_m$ has $V$ possible values $\{a_1, ..., a_v\}$, and when the dataset $D$ is partitioned using feature $F_m$, $V$ subsets are generated. The $v$-$th$ subset containing all the samples in $D$ with $a_v$ values on the feature $F_m$ in $D$, and is marked as $D^v$. $|D^v|$ denotes the number of samples when feature $F_m = a_v$. The smaller the $Gini\_index(D, F_m)$, the more the data uncertainty in the dataset is reduced by dividing by $F_m$, the more important $F_m$ is for the result. Assuming that the random forest consists of t trees, then the importance score of Fm can be calculated using equation (3):

$$I(F_m) = \frac{\sum_{i=1}^{t} Gini\_index(D_i, F_i)}{\sum_{m=1}^{M} \sum_{i=1}^{t} Gini\_index(D_i, F_i)} \qquad (3)$$

However, the data set is very small. Due to the sample division and the selection of the corresponding features of the random forest are all random, the prediction accuracy rate obtained from each time cross-validation will be different, and the difference may be very large. Only one Random Forests calculation is conducted, and there is experimental contingency. Therefore, we need to repeat $R$ ($R > 50$ generally) times of random forest feature evaluation, and each time there will be a score $I_r(F_m)$ of feature $F_m$ and a prediction accuracy $G_r$ obtained from cross-validation. The combined importance score for feature $F_m$ can be expressed as:

$$I(F_m) = \frac{\sum_{r=1}^{R} [I_r(F_m), G_r]}{\sum_{m=1}^{M} \sum_{r=1}^{R} [I_r(F_m), G_r]} \qquad (4)$$

After obtaining the scores for each feature, rank them from largest to smallest. Because the scores have been normalized, the importance of each feature can be clearly compared. This paper plans to select the top half of the features as important features for SISSO calculation.

**2 SISSO algorithm for symbolic regression**

Once the important features have been initially screened, SISSO can be used to filter the key features. To give a non-linear relationship between the material property "P" and the feature "f", the feature vector f is mapped in a higher dimensional space by a function $H: \mathbb{R}^{M_1} \to \mathbb{R}^{M_2}$. This is done by using the starting points $\phi_0'$, which indicate the important features screened from random forest, and then recursively performing algebraic combination of algebraic operations on these starting features to expand the features space. The operators set can be defined as:

$$H^{(m)} \equiv \{I, +, -, \times, \div, exp, log, |-|, \sqrt{}, \char`\^(-1), \char`\^2, \char`\^3\}[\varphi_1, \varphi_2] \qquad (5)$$

where $\varphi_1$ and $\varphi_2$ are components in the features space, and the superscript (m) indicates that a dimensional analysis is performed to preserve only meaningful combinations. In each iteration, $H^{(m)}$ will operate on all available combinations, and the features space will grow recursively, resulting in the features space $\phi$:

$$\phi \equiv \cup_{q=1}^{Q} \widehat{H}^{(m)}[\varphi_1, \varphi_2], \forall \varphi_1, \varphi_2 \in \phi_{q-1} \qquad (6)$$

The quantity of elements in $\phi$ grows quickly with the number of iterations, and the number of candidate features $M$ also increases rapidly. In order to find the key features affecting $P$, using the principle of *CS*, an attempt is made to find $c \in \mathbb{R}^M$ with the least non-zero terms satisfying $Fc = P$, which can be written as:

$$\underset{c:Fc=P}{\mathrm{argmin}} \|c\|_0 \tag{7}$$

where $P = (P_1, ..., P_N)^T$ is the output property column vector $F = \{f_1, ..., f_N\}^T$ is the input ($N \times M$) dimensional feature matrix, $\|c\|_0$ is the number of non-zero elements of vector $c$. $c$ is n-sparse, it means that $\|c\|_0 \leq n$. However, there is a significant disadvantage: when M is very large, that is when the number of reference features is very large, it is computationally infeasible because solving Eq. (7) is an NP-hard problem and only exhaustive methods can be used to find the solution $c$. Meanwhile, CS requires [3,4]: to restore a stable unique n-sparse $c$ from $F$ and $P$, it must also satisfy:

$$N \geq Cn \ln M \tag{8}$$

where C is a constant, usually between 1 and 10 [5]. This limits the number of features M to $e^{N/C_n}$ or less. But the huge features space obviously no longer meets the requirements of the effectiveness of CS.

To solve the problem of the large features space, SISSO introduced sure independence screening (SIS). It makes

$$\omega = PF^T \tag{9}$$

where $\omega = (\omega_1, ..., \omega_M)^T$ and $\boldsymbol{F} = \{\boldsymbol{F_1}, ..., \boldsymbol{F_M}\}$. If $\omega_i$ is larger, then $F_i$ is more strongly correlated with P. $|\omega_i|$ can be sorted from largest to smallest so that a features subspace $S_{1D}$ of an appropriate size can be selected, where $F_{1D}$ is the column of $F$ with the highest correlation with $P$. If you only want to find a relationship between the property and a combination of features, this is actually already obtained.

If one wishes to obtain $n(n > 1)$ dimensional descriptors, which means a relationship between the property defined in terms of $n$ sets of feature combinations. There also needs to define the residuals of the n-dimensional descriptors as:

$$\Delta_{nD} \equiv P - F_{nD}c_{nD} \tag{10}$$

where $F_{nD}$ is the column whose correlation with P is greatest in the $n$-$th$ SIS, $C_{nD} = F_{nD}^T F_{nD}^{-1} F_{nD}^T P$ is the least-squares solution of the $F_{nD}$ to $P$ fit. Then SIS selects the features subspace $S_{nD}$ which has the largest correlation with $\Delta_{(n-1)D}$ and combines $S_{nD}$ with $S_{(n-1)D}$ to form $S_n$. The size of the $S_n$ feature space is required to satisfy the requirement that a stable sparse solution $c$ can be solved.

Finally, CS is exploited in the feature space $S_n$ to obtain the best $n$-dimensional descriptor between a few key features and the property. The flow diagram of the overall algorithm is shown in the Fig.S1.

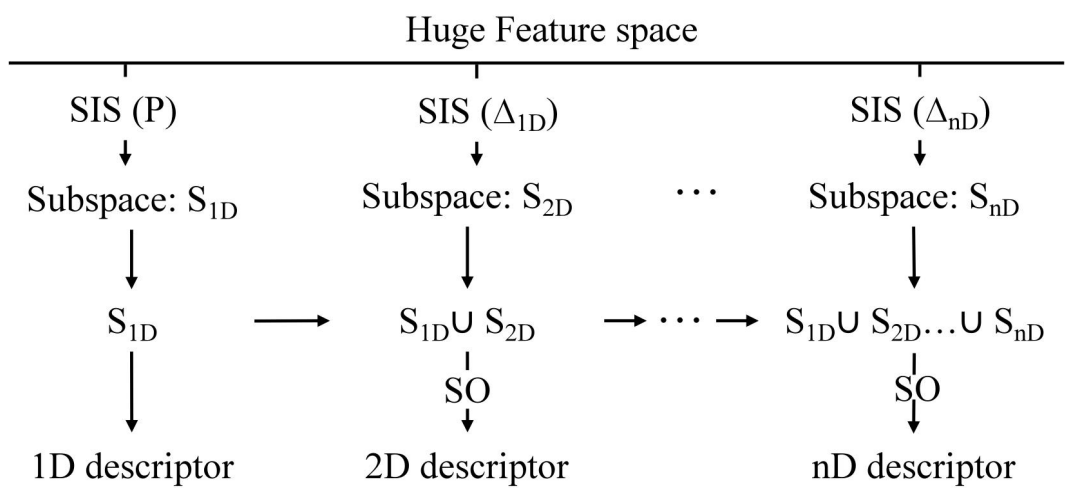

Fig.S1 Flowchart of SISSO algorithm.

## 3 RESULT

Table S1 Classification performance of Random Forests algorithm.

| | Random Forests (8 features) | | |
|---|---|---|---|
| Number of samples | Training_Accuracy | Testing_Accuracy | Training_Time |
| 45 | 1.000 | 0.910 | < 1 S |
| 75 | 1.000 | 0.911 | < 1 S |
| 150 | 0.980 | 0.933 | < 1 S |
| 224 | 0.978 | 0.893 | < 1 S |

Table S2 Two-dimensional descriptors derived by symbolic regression using SISSO and RF-SISSO in 224-sample subset.

| Datasets | | SISSO | RF-SISSO |
|---|---|---|---|
| 1 | d1 | $d_{AB} \times (V + nB) \times |IE_B - \chi_A/nA|$ | $[(\nu_A + \nu_B)/V] \times \exp(A - B)$ |
| | d2 | $(rcov_A/CN_A)/(nA/nB - \chi_A/\chi_B)$ | $\exp(V/\chi_B)(IE_B + |EA_B - \chi_A|)$ |
| 2 | d1 | $[(V/IE_B)/\chi_A]/(V + rcov_A/rcov_B)$ | $(V/\chi_A)^3 \times \nu_A/(\chi_B)^3$ |
| | d2 | $|d_{AB}/rcov_A| - (V/nA) \times (nB)^2$ | $(EA_A - 2IE_B - \chi_A)/(\chi_A/V)$ |
| 3 | d1 | $(\chi_A + \chi_B)/(\chi_A/V + nB \times \chi_A)$ | $V \times \chi_B \times [\exp(-\nu_A) + \exp(-\chi_A)]$ |
| | d2 | $|\chi_A/V| - (IE_A - \chi_B) \times |nA - nB|$ | $V \times (EA_A + IE_B) \times (|\nu_A - \nu_B| - \nu_A)$ |
| 4 | d1 | $d_{AB} \times (V + nB) \times (\chi_A/nA - IE_B)$ | $V \times \chi_B \times [\exp(-\nu_A) + \exp(-\chi_A)]$ |
| | d2 | $\left|\exp\left(-\frac{\chi_A}{\nu_B}\right) - \exp(-|CN_A - CN_B|)\right|$ | $(IE_B - 2\chi_B)/(\chi_A/V)$ |

Table S3 Two-dimensional descriptors derived by symbolic regression using SISSO and RF-SISSO in 150-sample subset.

| Datasets | | SISSO | RF-SISSO |
|---|---|---|---|
| 1 | d1 | $(V - nA)[IE_B/\chi_A - d_{AB}/rcov_B]$ | $[(\nu_A + \nu_B)/\chi_B] \times (\chi_A/V)^2$ |
| | d2 | $(CN_A \times d_{AB})^3 \times \exp(\chi_B/nB)$ | $(EA_A - IE_B - |IE_B - \chi_B|)/(\chi_A/V)$ |
| 2 | d1 | $V/\exp(CN_B) + \chi_A/(V \times \chi_B)$ | $(IE_B \times \chi_A)/((IE_B - \chi_B) \times V \times \chi_B)$ |
| | d2 | $(CN_A \times IE_B)^3 + |CN_A - V \times CN_B|$ | $V^9 \times \nu_A \times \exp(\nu_B)$ |
| 3 | d1 | $(\chi_A + \chi_B)/(\chi_A/V + nB \times \chi_A)$ | $V \times IE_B \times [\exp(-\nu_A) + \exp(-\chi_A)]$ |
| | d2 | $CN_A \times IE_A \times |nB \times CN_B - |CN_A - CN_B||$ | $||EA_B - \chi_A| - EA_B| - |2\chi_A - \chi_B|$ |
| 4 | d1 | $|\chi_B - IE_A/V| - |EA_B - \chi_B/V|$ | $(2EA_A + IE_B - EA_B)/(\chi_A/V)$ |
| | d2 | $(CN_B \times IE_B)^3 \times (EA_B \times \chi_A)/nA$ | $EA_B + \chi_A + (\chi_B)^3/(IE_B \times \chi_A)$ |

Table S4 Two-dimensional descriptors derived by symbolic regression using SISSO and RF-SISSO in 75-samples subset.

| Datasets | | SISSO | RF-SISSO |
|---|---|---|---|
| 1 | d1 | $(IE_B - \chi_A)/(nB \times |\chi_B - IE_A/V|)$ | $[(V \times \chi_B)^3 \times v_B]/(V \times \chi_A)^3$ |
| | d2 | $(V \times rcov_A)/nB \times (IE_B + \chi_B/nA)$ | $[(IE_B)^3 \times (v_A)^4]/(EA_A - EA_B)$ |
| 2 | d1 | $(V \times \chi_B - IE_A) \times \exp(CN_A/nB)$ | $[(IE_B \times \chi_B)^3 \times (V)^3]/(\chi_A)^3$ |
| | d2 | $(V \times \chi_B - IE_A) \times \exp(CN_A + CN_B)$ | $(V/\chi_A)^3 \times (\chi_B)^6$ |
| 3 | d1 | $(V)^2 \times \chi_B \times [(IE_B \times d_{AB})^3$ | $(V \times IE_B)^3 \times (EA_B + \chi_B)/\chi_A$ |
| | d2 | $(V)^2 \times IE_B \times [(IE_B \times d_{AB})^3$ | $(V/\chi_A)(IE_B \times \chi_B)^3$ |
| 4 | d1 | $(\chi_A/nA - IE_B) \times (V \times d_{AB} + rcov_A)$ | $V \times IE_B \times [\exp(-v_A) + \exp(-\chi_A)]$ |
| | d2 | $V \times nA \times \chi_B \times (d_{AB} \times IE_B)^2$ | $[V \times (IE_B + \chi_B)]/\exp(V/v_B)$ |

Table S5 Two-dimensional descriptors derived by symbolic regression using SISSO and RF-SISSO in 45-sample subset.

| Datasets | | SISSO | RF-SISSO |
|---|---|---|---|
| 1 | d1 | $(V \times \chi_B)^3 \times \exp(IE_A/\chi_A)$ | $[(V \times \chi_B)^3 \times IE_B]/v_A$ |
| | d2 | $\exp(IE_A/\chi_A + V \times \chi_B)$ | $(IE_B \times \chi_B \times V^2)/\chi_A^2$ |
| 2 | d1 | $(V \times \chi_B - IE_A) \times \exp(CN_A/nB)$ | $(V/\chi_A)^3 \times \chi_B^6$ |
| | d2 | $[\chi_A/(V \times nB)] \times \exp(CN_B/nA)$ | $(V/\chi_A)^3 \times (EA_A \times \chi_B)^6$ |
| 3 | d1 | $[\chi_A/(V \times \chi_B)] \times \exp(CN_B/nA)$ | $(V)^3 \times IE_B \times (\chi_A - \chi_B)$ |
| | d2 | $(V \times IE_B)^3 \times CN_B \times (\chi_A - \chi_B)$ | $IE_B \times \chi_B \times (V/\chi_A)^3$ |
| 4 | d1 | $\chi_B \times (V - nB) \times (d_{AB} \times IE_B)^3$ | $V/(EA_B/IE_B + \chi_A/\chi_B)$ |
| | d2 | $(d_{AB} \times IE_B)^3 \times (V \times \chi_B)/CN_A$ | $(v_A)^3/(EA_A + EA_B - \chi_A)$ |

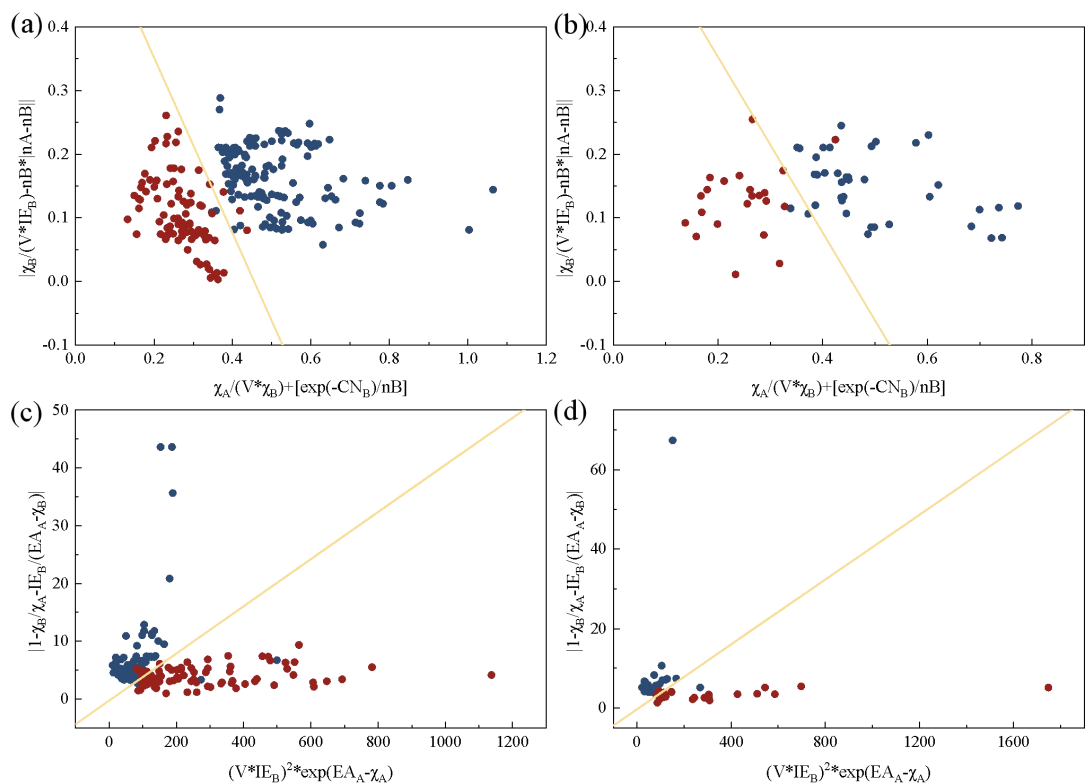

Fig.S2 Visualizations of SVC classification for 2D descriptors based on SISSO training in a 224-sample subset. (a) training and (b) testing sets with operators from SISSO; (c) training and (d) testing sets with operators from RF-SISSO.

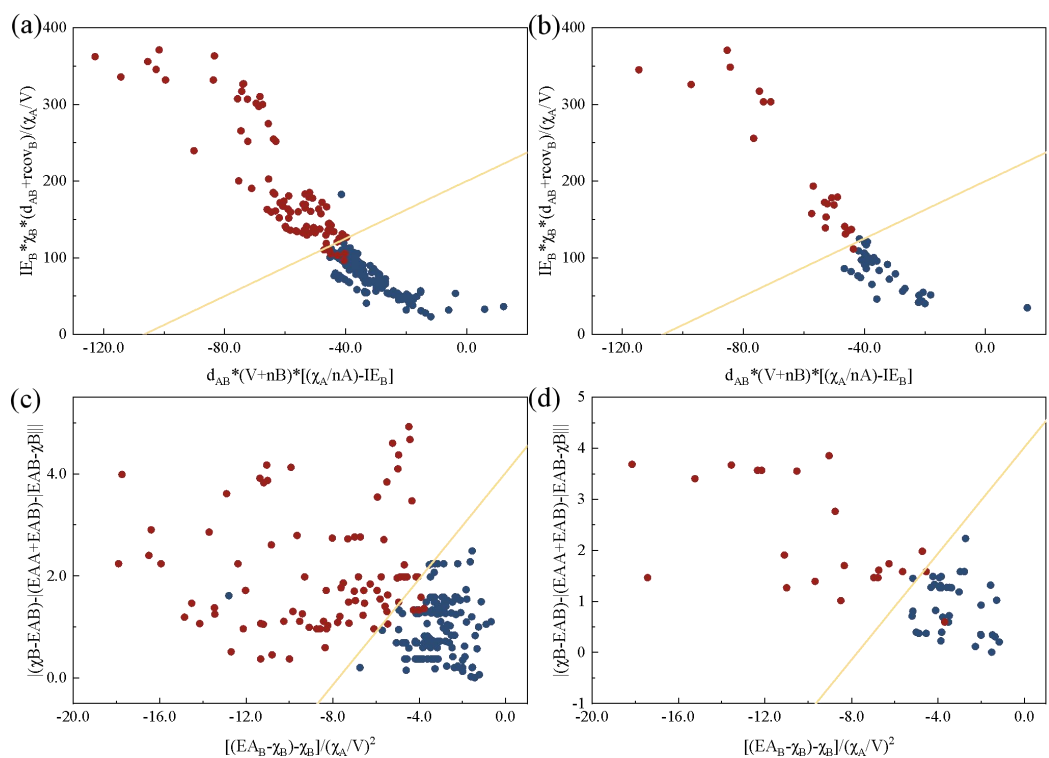

Fig.S3 SVC classification visualizations for 2D descriptors based on SISSO training in a 150-sample subset. (a) training and (b) testing sets with operators from SISSO; (c) training and (d) testing sets with operators from RF-SISSO.

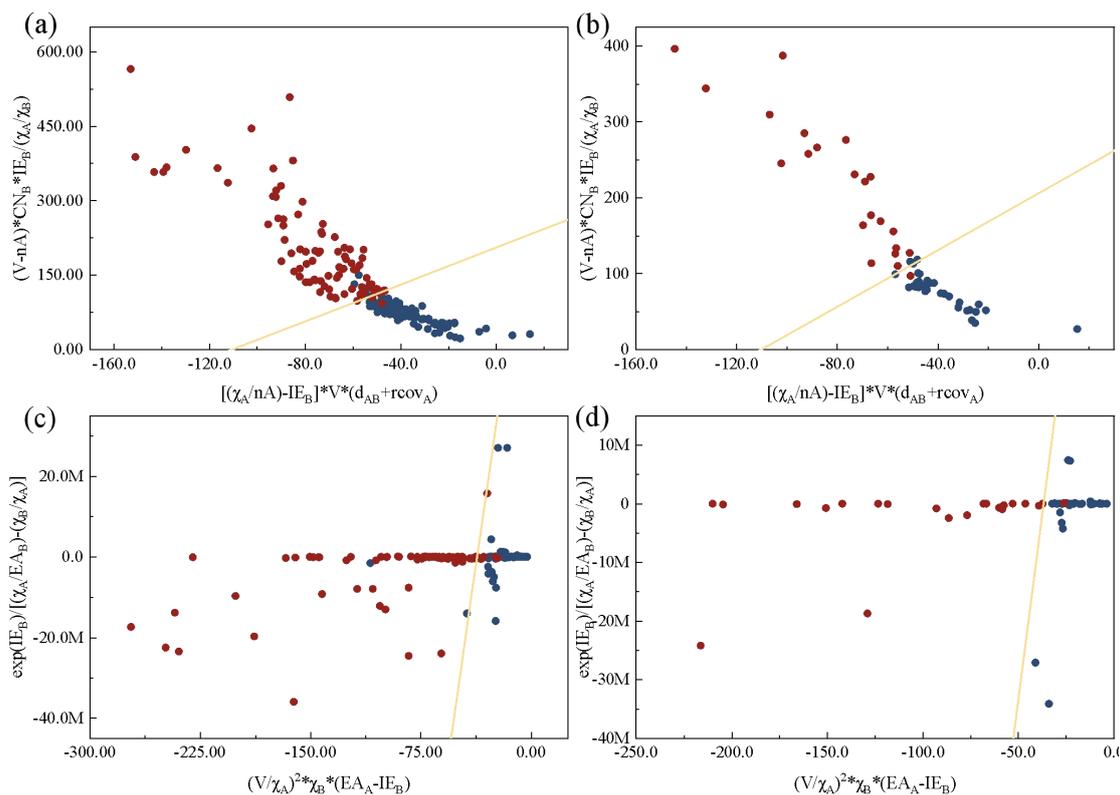

Fig.S4 SVC classification visualizations for 2D descriptors based on SISSO training in a 75-sample subset. (a) training and (b) testing sets with operators from SISSO; (c) training and (d) testing sets with operators from RF-SISSO.

Table S6 Comparison of regression efficiency and accuracy between RF-SISSO and SISSO on small sample sizes

| System | Algorithm | r_train | RMSE_train | Time (s) | r_test | RMSE_test |
|---|---|---|---|---|---|---|
| PFC-20 | RF-SISSO | 0.999 | 0.001 | 5417.28 | 0.817 | 0.022 |
|  | SISSO | 0.999 | 0.001 | 15454.56 | 0.847 | 0.012 |
| PFC-40 | RF-SISSO | 0.999 | 0.002 | 599.6 | 0.972 | 0.006 |
|  | SISSO | 0.999 | 0.002 | 11062.56 | 0.849 | 0.013 |
| PO3-20 | RF-SISSO | 0.998 | 0.007 | 2686.08 | 0.962 | 0.080 |
|  | SISSO | 0.997 | 0.007 | 10811.44 | 0.883 | 0.079 |
| PO3-40 | RF-SISSO | 0.998 | 0.010 | 6616.8 | 0.948 | 0.066 |
|  | SISSO | 0.998 | 0.010 | 15885.6 | 0.885 | 0.110 |

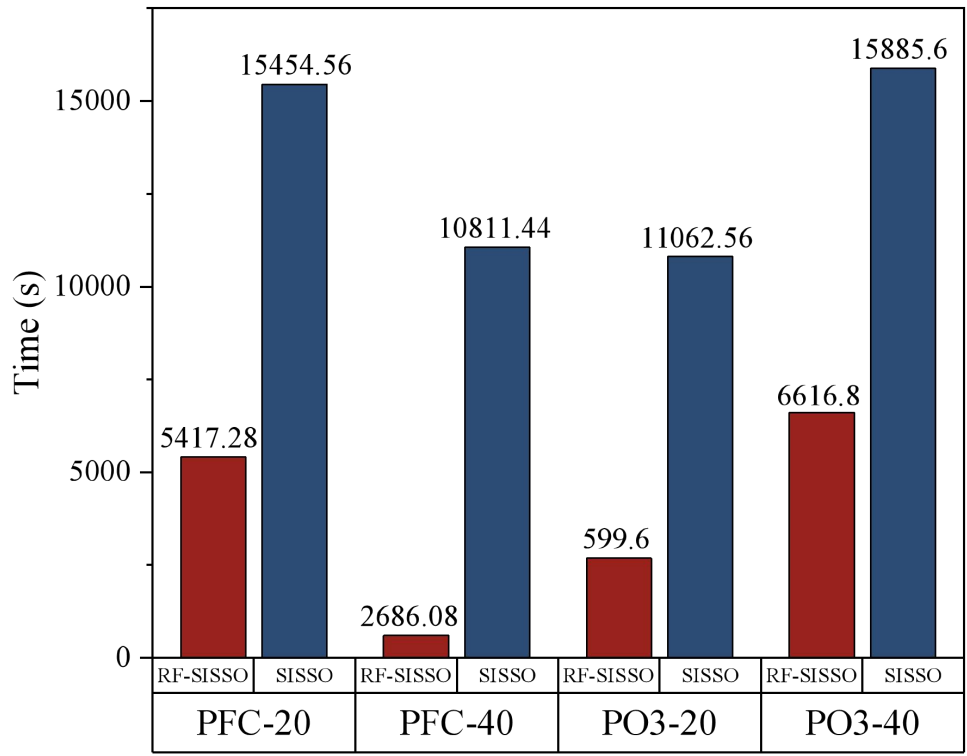

Fig.S5 Regression time of SISSO and RFSISSO for different sample numbers in PFC and PO3 system.